\documentclass[10pt,twocolumn,letterpaper]{article}

\usepackage{authblk}
\usepackage{wacv}    
\usepackage[accsupp]{axessibility}
\usepackage{graphicx}
\usepackage{amsmath}
\usepackage{amssymb}
\usepackage{booktabs}
\usepackage{multirow}
\include{macros}

\newcommand\blfootnote[1]{%
  \begingroup
  \renewcommand\thefootnote{}\footnote{#1}%
  \addtocounter{footnote}{-1}%
  \endgroup
}
\usepackage[capitalize]{cleveref}
\crefname{section}{Sec.}{Secs.}
\Crefname{section}{Section}{Sections}
\Crefname{table}{Table}{Tables}
\crefname{table}{Tab.}{Tabs.}

\begin{document}

\title{Dynamic Appearance Modeling of Clothed 3D Human Avatars \\ using a Single Camera}
\author{Hansol Lee\textsuperscript{1} \quad
Junuk Cha \textsuperscript{1} \quad
Yunhoe Ku \textsuperscript{1} \quad
Jae Shin Yoon\textsuperscript{2*} \quad
Seungryul Baek\textsuperscript{1*}
}
\affil{\textsuperscript{1}UNIST \qquad \textsuperscript{2}Adobe Research}

\twocolumn[{
\maketitle
\begin{center}
    \captionsetup{type=figure}
    \includegraphics[width=1\textwidth]{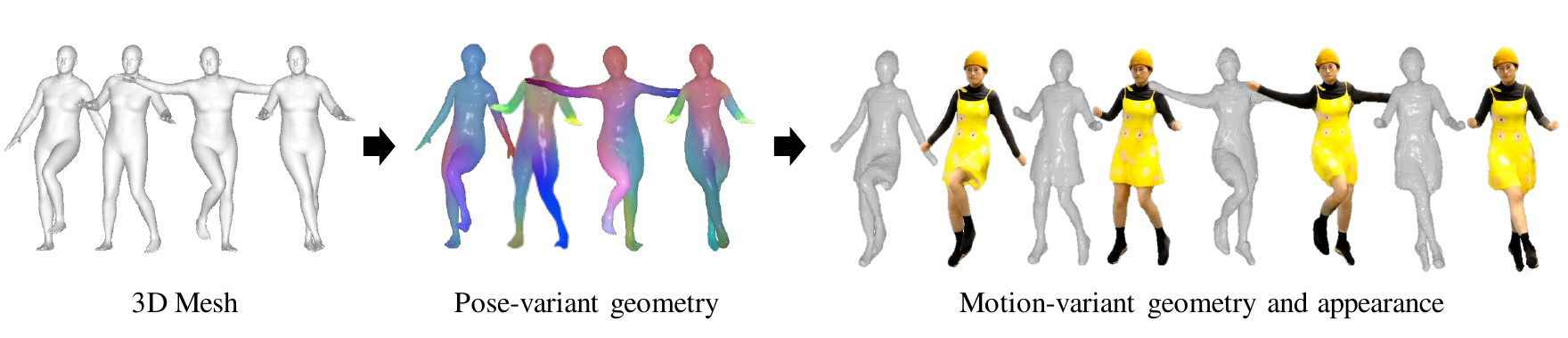}
    \captionof{figure}{The clothed 3D human avatars with motion-variant appearance and its intermediate representation. Using a 3D model mesh, we first estimate explicit pose-variant geometry. We then reconstruct implicit motion-dependent clothed 3D human avatars conditioned by the temporal surface correspondences from the explicit geometry. The color coding in the pose-variant geometry represents a visual interpretation of our temporal modeling results, demonstrating 3D displacement over time.}
    \vspace{2mm}
    \label{fig:teaser}
\end{center}
}]

\blfootnote{Co-corresponding authors$*$.}

\begin{abstract}
   The appearance of a human in clothing is driven not only by the pose but also by its temporal context, i.e., motion. However, such context has been largely neglected by existing monocular human modeling methods whose neural networks often struggle to learn a video of a person with large dynamics due to the motion ambiguity, i.e., there exist numerous geometric configurations of clothes that are dependent on the context of motion even for the same pose. In this paper, we introduce a method for high-quality modeling of clothed 3D human avatars using a video of a person with dynamic movements. The main challenge comes from the lack of 3D ground truth data of geometry and its temporal correspondences. We address this challenge by introducing a novel compositional human modeling framework that takes advantage of both explicit and implicit human modeling. For explicit modeling, a neural network learns to generate point-wise shape residuals and appearance features of a 3D body model by comparing its 2D rendering results and the original images. This explicit model allows for the reconstruction of discriminative 3D motion features from UV space by encoding their temporal correspondences. For implicit modeling, an implicit network combines the appearance and 3D motion features to decode high-fidelity clothed 3D human avatars with motion-dependent geometry and texture. The experiments show that our method can generate a large variation of secondary motion in a physically plausible way. 
\end{abstract}

\section{Introduction}
\label{sec:intro}
The modeling of 3D clothed human avatars with a realistic secondary motion enables various applications such as physically plausible film-making, gaming, augmented reality (AR), virtual reality (VR), and e-commerce industries. Existing methods~\cite{Zhao_2022_CVPR,alldieck2018video,alldieck2019tex2shape,Jiang_2022_CVPR,chen2021animatable} have studied how to model appearance and geometry of a human using a single camera for the widespread of such self-avatars in those applications. However, many of them have largely ignored the importance of motion-dependent appearance modeling for the realistic animation of a clothed 3D human avatar. In this paper, we model dynamic humans, a clothed 3D human avatar with high-fidelity motion-dependent geometry and texture, using a single camera.

Modeling motion-dependent appearance is challenging due to the lack of 3D ground truth data. In particular, we are missing three types of data: 1) 3D body pose that aligns with images to encode the pose-to-appearance relationship; 2) 3D surface to learn to reconstruct high-quality 3D geometry; and 3) temporal surface correspondences to model motion-dependent geometry. While the recent advance in 3D pose prediction allows us to have accurate 3D pose, obtaining high-quality 3D temporal correspondences and geometry is still challenging, which prevents us from modeling the motion-dependent appearance of a high-fidelity clothed 3D human avatar.  

Existing monocular 3D human modeling has addressed these challenges in two ways: explicit and implicit modeling. For explicit approach, previous works~\cite{Zhao_2022_CVPR,alldieck2018video, alldieck2019tex2shape,bhatnagar2019multi} have mainly focused on the reconstruction of high-fidelity static geometry, e.g., the surface for a A-posed person, by predicting the person-specific residual geometry of an explicit 3D body model, e.g., SMPL~\cite{Loper_2015_TOG} which provides topology-invariant mesh surface. While such explicit representation provides explicit correspondences over different body poses, this representation significantly limits the capability of the residual geometry to describe its local details due to the nature of the fixed topology of the 3D body model. 

The implicit approaches~\cite{xu2021h, Saito_2020_CVPR, Xiu_2022_CVPR, saito2019pifu, Hong_2021_CVPR, Peng_2021_CVPR, Weng_2022_CVPR} effectively overcome the limitation of explicit methods by utilizing a number of variations of implicit representations such as occupancy~\cite{mescheder2019occupancy}, signed/unsigned distance fields~\cite{park2019deepsdf,chibane2020neural} or radiance fields~\cite{mildenhall2020nerf} where they are more flexible to describe a complex and challenging shape since they are free from a fixed topology. However, these implicit models inhibit the establishment of the temporal correspondences of the reconstructed 3D surface.

In this paper, we use the complementary properties of the explicit and implicit representations by newly designing a compositional human modeling framework. For explicit modeling, we perform joint modeling of 3D body pose and pose-dependent geometry of humans by predicting explicit geometric offsets between the clothed humans (observed from a video) and unclothed parametric 3D body model (i.e., SMPL~\cite{Loper_2015_TOG}). This explicit model is topology-invariant to body poses so that it enables the joint training of the appearance and 3D surface motion features in the canonical UV coordinates. The modeled appearance and motion features are conditioned in two implicit functions dedicated to decoding motion-dependent texture and geometry, respectively. We highlight that securing well-aligned explicit geometry is important to establish accurate  temporal surface correspondences, which allows the implicit models to reconstruct the high-fidelity dynamic human avatars with plausible local details. 

The experiments show that our method can generate motion-dependent avatars with plausible clothes movements even for unseen pose sequences. We compare our method with state-of-the-art avatar reconstruction methods and it outperforms other methods in generating motion-dependent avatars with plausible clothes movements, thereby highlighting the effectiveness of our compositional framework.

Our contribution can be summarized as follows:
\begin{itemize}
    \item We introduce a novel compositional framework that can take advantage of both implicit and explicit representation to model the high-fidelity clothed 3D human avatar with motion-dependent appearance (texture and geometry).   
    \item We demonstrate that conditioning motion information on implicit functions enables more effective learning for the video with a person with large dynamics and capturing the subtle variations of motion-dependent appearance, leading to physically plausible human avatar animation.
    \item We demonstrate that our method is capable of generating more realistic renderings of dynamic human appearance, particularly for large motions, compared to other methods. Moreover, our approach exhibits strong generalizability to unseen poses, indicating its effectiveness in handling variations in pose and motion.
\end{itemize}

\begin{figure*}
    \centering
    \includegraphics[width=1\linewidth]{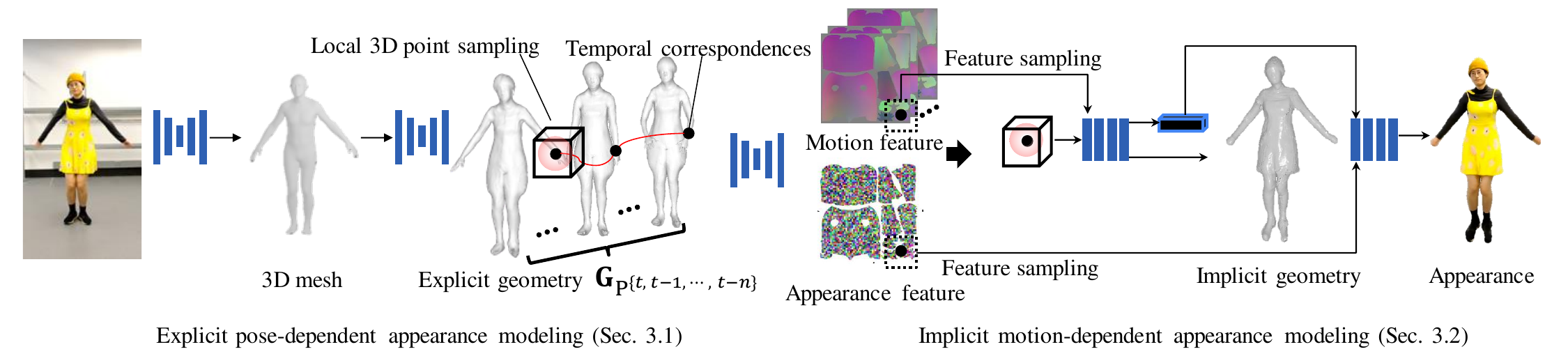}
    \caption{{\textbf{The system overview}. We propose a compositional framework that combines explicit and implicit representations of clothed 3D human avatars. First, we reconstruct explicit geometry by predicting the surface residuals from a 3D body mesh model. Using their temporal correspondences, we extract high-dimensional motion and appearance features, which are conditioned on an implicit network to reconstruct the clothed 3D human avatar with motion-dependent geometry and appearance.}}
    \vspace{1mm}
    \label{fig:pipeline}
\end{figure*}

\section{Related works}
\label{sec:RelatedWorks}
\noindent\textbf{3D Clothed Human Reconstruction.} Existing methods~\cite{Kanazawa_2018_CVPR,Kocabas_2021_ICCV,Choutas_2020_ECCV} utilize parametric 3D body models (e.g., SMPL and SMPL-X~\cite{Loper_2015_TOG,Pavlakos_2019_CVPR}) to represent 3D human pose and shape. However, these models are fundamentally limited in expressing fine details such as clothing since they were designed specifically for minimally clothed humans. To overcome this limitation, other methods~\cite{alldieck2018detailed,alldieck2019learning,alldieck2019tex2shape,lazova2019360,zhu2019detailed,xiang2020monoclothcap,xiu2022econ} estimate 3D offsets on top of body mesh vertices. While these methods add some details, their performance on loose clothing is limited due to the constrained 3D mesh topologies. Researchers have proposed various implicit representation techniques for human reconstruction and modeling to address these limitations, including signed distance fields~\cite{MetaAvatar:NeurIPS:2021}, occupancy fields~\cite{Saito_2020_CVPR, Hong_2021_CVPR, Xiu_2022_CVPR, saito2019pifu, Peng_2021_CVPR}, and NeRF-based methods~\cite{xu2021h, Weng_2022_CVPR}. ICON~\cite{Xiu_2022_CVPR} leverages local features to model the high-frequency details of the clothed humans' local geometry. ECON~\cite{xiu2022econ} improves upon ICON's limited generalization ability for various clothing topologies, particularly loose clothing, by completing the full 3D shape using front and back normals and surface information extracted from RGB images. 
HumanNeRF~\cite{Weng_2022_CVPR} reconstructs clothed avatars as a free-viewpoint mixture of affine fields. Despite these advances, such approaches often suffer from temporal inconsistency in shape reconstruction because of the lack of temporal surface correspondences, leading to physically implausible 3D animations.

\noindent\textbf{Animatable Human Avatars.}
Some works~\cite{xu2021h, Jiang_2022_CVPR} proposed methods for the co-training of the SDF and NeRF network constraining neural radiance field by implicit human body model. However, they require a large corpus of posed human scans and can not generalize well to the poses beyond the distribution of the training set.
SCANimate~\cite{Saito_2021_CVPR} introduced the end-to-end trainable framework to build a high-quality parametric clothed human model from raw scans by encoding locally pose-aware implicit surface representation to model pose-dependent clothing deformation. However, it requires massive 3D ground truth data, which is not possible to obtain using a single camera.
Animatable NeRF~\cite{Peng_2021_ICCV} designed a novel transformation field between view and canonical space to better memorize the seen poses. Zheng~\textit{\etal}~\cite{Zheng_2022_CVPR} proposed a method to use a set of structured local radiance fields attached to a human body template. 
HF-Avatar~\cite{Zhao_2022_CVPR} proposed coarse-to-fine explicit representation using a dynamic surface network that refines the unclothed body shape by estimating clothed shape offsets. Anim-NeRF~\cite{chen2021animatable} proposed a geometry-guided deformable NeRF for explicit pose-guided deformation that enables explicit control of a human avatar. While those existing methods have shown promising results, they often struggle to learn a dynamic avatar with large movements due to the absence of explicit conditioning temporal analysis to model motion-dependent appearance.


\section{Method}
\label{sec:method_overview}
Our method aims to generate high-fidelity clothed 3D human avatars with motion-dependent appearance from a single camera. It consists of two main components: explicit pose-dependent appearance modeling (Sec.~\ref{sec:method_explicit}) and implicit motion-dependent appearance modeling (Sec.~\ref{sec:method_implicit}). The explicit modeling component generates pose-dependent geometry and appearance from a 3D body model whose temporal information is used to extract motion features. The implicit modeling component decodes the motion features to reconstruct motion-dependent appearance and geometry. Figure~\ref{fig:pipeline} provides an overview of our pipeline.

\begin{figure*}
    \centering
    \includegraphics[width=1\linewidth]{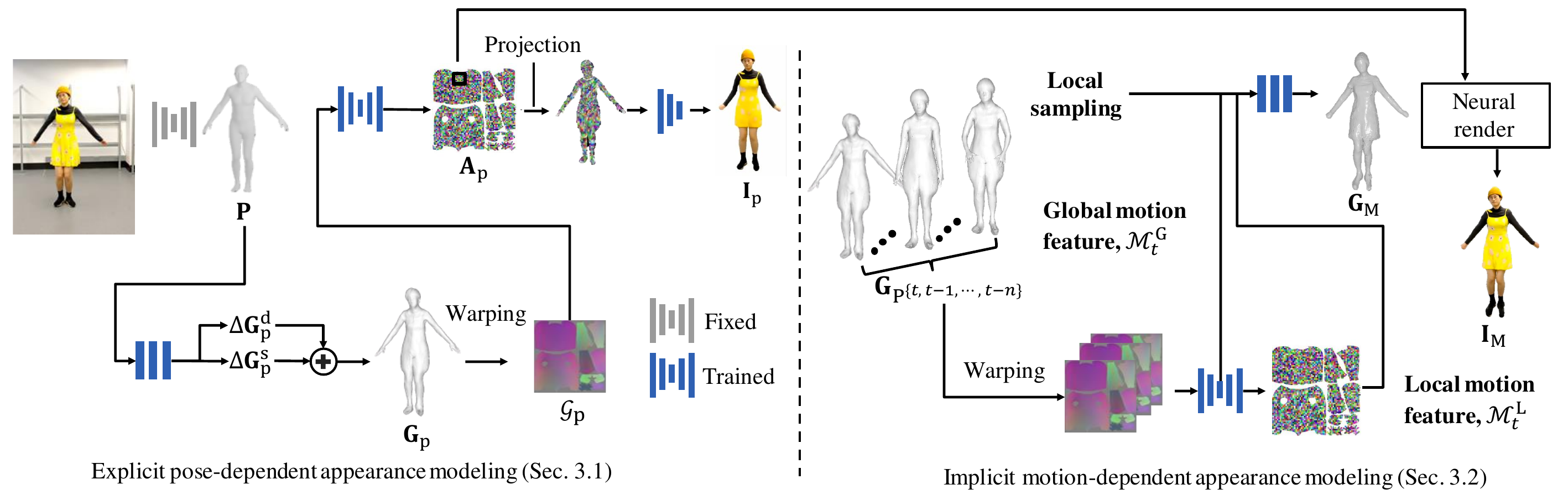}
    \caption{{(Left) Explicit pose-dependent appearance modeling with 3D performance capture. It predicts the pose-dependent geometry $\mathbf{G}_{\text{p}}$  and appearance feature $\mathbf{A}_{\text{p}}$ on single image. Implicit motion-dependent appearance modeling. (Right) It predicts the motion-conditioned clothed human mesh $\mathbf{G}_{\text{M}}$ under the target image pose. The global and local motion features are extracted from the previous $n$ frame's explicit geometries.}}
    \vspace{1mm}
    \label{fig:pipeline_stage1}
\end{figure*}

\subsection{Explicit Pose-dependent Appearance Modeling.}
\label{sec:method_explicit}
In this section, we model the person-specific 3D human avatar in clothing that captures the pose-dependent appearance and shape:
\begin{eqnarray}
\centering
\mathbf{G}_{\text{p}}, \mathbf{A}_{\text{p}} \leftarrow f(\mathbf{P}).
\label{eq:1}
\end{eqnarray}
$f$ is the modeling function that generates pose-dependent geometry $\mathbf{G}_{\text{p}}\in\mathbb{R}^{n\times3}$ and appearance feature $\mathbf{A}_{\text{p}}\in\mathbb{R}^{u\times v\times 64}$, where $n$ is the number of vertices, and $\mathbf{A}$ is the canonical UV appearance map whose width and height are $u$ and $v$, respectively. $\mathbf{P}\in\mathbb{R}^{n\times3}$ is the vertices of 3D mesh model (i.e., SMPL~\cite{Loper_2015_TOG}) parameterized by a human pose parameter ${\text{p}}$.

Inspired by existing human geometry processing~\cite{ma2021scale}, we design the modeling function $f$ in a way that predicts the pose-dependent geometry in the residual space for more effective generalization to unseen body poses as follows:
\begin{eqnarray}
\centering
\mathbf{G}_{\text{p}}, \mathbf{A}_{\text{p}} \leftarrow f(\mathbf{P}), \ \ \ \ \ \text{where} \ \ \ \ \mathbf{G}_{\text{p}}=\mathbf{P}+\Delta\mathbf{G}_{\text{p}}.
\label{eq:2}
\end{eqnarray}
$\Delta\mathbf{G}_{\text{p}}\in\mathbb{R}^{n\times 3}$ is the pose corrective residual.

We train the explicit modeling function $f$ using the following objectives:
\begin{eqnarray}
&&L_{\text{f}} = \omega_{\text{mask}} L_{\text{mask}} + \omega_{\text{normal}} L_{\text{normal}} + \omega_{\text{lap}} L_{\text{lap}}  \nonumber \\ 
&&+ \omega_{\text{rgb}}L_{\text{rgb}} +  \omega_{\text{vgg}}L_{\text{vgg}} + \omega_{\text{GAN}}L_{\text{GAN}},
\label{eq:3}
\end{eqnarray} 
where mask loss $L_\text{mask}$ encourages the silhouette of the reconstructed 3D mesh to be aligned with the ground truth mask by measuring their difference: $L_\text{mask}= \|\mathbf{M}_{\rm p}-\mathbf{M}_{\rm gt}\|$ where $\mathbf{M}_{\rm p}$ is the rendered 2D mask from the reconstructed geometry $\mathbf{G}_{\rm p}$, and $\mathbf{M}_{\rm gt}$ is the ground-truth mask detected from the video using existing human segmentation method~\cite{gong2018instance}.
In a similar way, the normal loss $L_\text{normal}$ penalizes the difference between the rendered surface normal $\mathbf{N}_{\rm p}$ from $\mathbf{G}_{\rm p}$ and the detected one $\mathbf{N}_{\rm gt}$ from the existing surface normal detection method~\cite{jafarian2021learning}, i.e.,  $L_\text{normal}= \|\mathbf{N}_{\rm p}-\mathbf{N}_{\rm gt}\|$, which leads to the reconstruction of the detailed geometry. 
The Laplacian smoothing loss $L_\text{lap}$ regularizes the distribution of the predicted offsets by minimizing the second-order derivatives of the 3D surface similar to the approach from Grassal et al.~\cite{grassal2022neural}. RGB loss $L_\text{rgb}$ measures the L1 difference between the rendered RGB image and the ground truth RGB image, i.e., $L_\text{rgb}= \|\mathbf{I}_{\rm p}-\mathbf{I}_{\rm gt}\|$, where $\mathbf{I}_{\rm p}$ is the rendered image and $\mathbf{I}_{\rm gt}$ is the ground truth image from the video. Additionally, we use VGG loss $L_\text{vgg}$~\cite{simonyan2014very} to consider the content and style of the rendered image. Finally, we apply the unconditional adversarial loss $L_\text{GAN}$~\cite{demir2018patch} for producing realistic human texture where the rendered image is used as a fake and the ground truth as real. \\
The overview of the network for this section is shown on the left side of Fig.~\ref{fig:pipeline_stage1}, and please also refer to the supplementary material for the network details.

\noindent\textbf{Implementation Details.} In this section, our network inputs are SMPL pose parameters and outputs are pose-dependent human specific geometry offsets and appearance features.
\noindent We first estimate the SMPL pose parameters $\text{p}$ and camera pose $\mathbf{C}$ from a single video using a two-step approach. In the first step, we obtain the initial pose estimate using an existing 3D pose estimation method, e.g., VIBE~\cite{kocabas2020vibe}. In the second step, we optimize the initial pose based on the video cue by minimizing the difference of the 3D mesh surface with the detected surface (from an existing human surface detection, i.e., DensePose~\cite{guler2018densepose}) in the image space. From an RGB image, we obtain the 3D mesh $\mathbf{P}$, its corresponding $\rm p$, and $\mathbf{C}$. We use the VIBE~\cite{kocabas2020vibe} model for initial $\mathbf{P}$ and $\mathbf{C}$ estimates, and DensePose for obtaining surface correspondences with the 3D body model in the image space. We fix the shape of the 3D body model $\mathbf{P}$ with the mean shape. To better express mesh deformation, we subdivide each triangular face of the body mesh into four using edge subdivision. We up-sample the template body mesh vertices from 6,890 to 27,554. 

We enable the explicit appearance modeling function $f$, using two networks for geometry and appearance.

\noindent Given an unclothed 3D body mesh model~\cite{Loper_2015_TOG}, our approach is designed to generate pose-dependent geometric offset and appearance to reconstruct a person-specific avatar by following ~\cite{Grassal_2022_CVPR}.

\noindent The geometry network learns to predict the pose-dependent geometry offset. In practice, it consists of both static and dynamic offset, i.e., $\Delta\mathbf{G}_{\text{p}} = \Delta\mathbf{G}^\text{\rm s}_{\text{p}} + \Delta\mathbf{G}^{\rm d}_{\text{p}}$. While the static offset is a subject-specific offset across all frames, the dynamic offset is a pose-specific offset unique to each frame. The static offset $\Delta\mathbf{G}^\text{\rm s}_{\text{p}}$ is obtained by feeding a zero-value vector into the network. This vector has the exact dimensions as the SMPL pose parameters. The output serves as the static offset, establishing the subject's base shape. For the dynamic offset $ \Delta\mathbf{G}^{\rm d}_{\text{p}}$, the network uses SMPL pose parameters to generate pose-variant offsets. Static and dynamic offsets are combined during the training process to generate a comprehensive pose-dependent geometry offset. This architecture enables the model to capture the subject's inherent, pose-invariant geometry while also facilitating pose-variant adjustments.

\noindent The appearance networks is composed of two parts: appearance feature extraction and appearance texture rendering.
We employ a U-Net architecture for appearance feature extraction. The input to this U-Net is the geometry represented in UV coordinates, denoted as $\mathcal{G}_\text{p}=u^{-1}(\mathbf{G}_\text{p})$. The output is a pose-dependent appearance feature $\mathbf{A}_{\text{p}}$ in the UV coordinates. We warp these appearance features to the image space to render the texture using a final CNN layer. 

\subsection{Implicit Motion-dependent Appearance Modeling.}
\label{sec:method_implicit}
In this section, we model an implicit avatar with a motion-dependent appearance by encoding the temporal context of the explicit pose-dependent geometry $\mathbf{G}_{\text{p}}$ and appearance feature $\mathbf{A}_{\text{p}}$ modeled in Sec~\ref{sec:method_explicit}:
\begin{eqnarray}
\centering
\mathbf{o},\mathbf{c}\leftarrow g(\{\mathbf{G}_{\text{p}^{t}}, \mathbf{G}_{\text{p}^{t-1}}, \cdots, \mathbf{G}_{\text{p}^{t-n}}\},\mathbf{A}_{\text{p}^{t}}; \mathbf{X}),
\label{eq:4}
\end{eqnarray}
where $g$ is the implicit function designed with multi-layer perceptrons (MLPs) which take the sequence of 3D pose-dependent geometry $\{\mathbf{G}_{\text{p}^{t}}, \mathbf{G}_{\text{p}^{t-1}}, ..., \mathbf{G}_{\text{p}^{t-n}}\}$, explicit appearance feature $\mathbf{A}_{\text{p}^{t}}$ at the target pose, and the sampled 3D point $\mathbf{X}$; and outputs implicit avatar with the motion-dependent appearance in the form of occupancy $\mathbf{o}\in[0,1]$ and color $\mathbf{c}\in\mathbb{R}^{3}$. 

We further extract the high-dimensional motion features using an encoder $E_{\rm m}$ which allows encoding more discriminative and robust motion context~\cite{Yoon_2022_CVPR}:
\begin{eqnarray}
\centering
&&\mathbf{o},\mathbf{c}\leftarrow g(\{\mathcal{M}_{t}^{\rm G},\mathcal{M}_{t}^{\rm L}\},\mathbf{A}_{\text{p}^{t}}; \mathbf{X}) \nonumber \\
&&=g(E_{\rm m}\{\mathbf{G}_{\text{p}^{t}}, \mathbf{G}_{\text{p}^{t-1}}, \cdots, \mathbf{G}_{\text{p}^{t-n}}\},\mathbf{A}_{\text{p}^{t}}; \mathbf{X}).
\label{eq:4}
\end{eqnarray}
In practice, the motion encoder $E_{\rm m}$ extracts the high-dimensional features in the UV coordinates, which result in the high-dimensional global motion features $\mathcal{M}_{t}^{\rm G}\in\mathbb{R}^{1\times64}$ that guides the motion with respect to the entire body; and the high-dimensional local motion features $\mathcal{M}_{t}^{\rm L}\in\mathbb{R}^{u\times v\times64}$ that guides the motion of the local surface, e.g., secondary clothing movement as shown in Fig.~\ref{fig:pipeline_stage1}-(right). 


We train implicit modeling function $g$ using the following objectives:
\begin{eqnarray}
L_{\text{g}} = \omega_{\text{IOU}} L_{\text{IOU}} + \omega_{\text{color}} L_{\text{color}} + \omega_{\text{norm}} L_{\text{norm}} + \omega_{\text{eik}}L_{\text{eik}},
\label{eq:10}
\end{eqnarray}
where the IOU loss $L_\text{IOU}$ encourages the alignment between the estimated segmentation mask and the ground-truth mask. It is computed as the Intersection Over Union: $L_{\text{IOU}} = 1 - \text{IOU}(\mathbf{M}_{\text{M}}, \mathbf{M}_{\rm gt})$, where $\mathbf{M}_{\text{M}}$ is the rendered 2D mask from geometry $\mathbf{G}_\text{M}$ using a differentiable renderer, as described by~\cite{wiles2020synsin}. 

The color loss $L_{{\text{color}}}$ measures the L1 distance between the rendered color $\hat{\mathbf{c}}_{\mathbf{p}}$ at pixel $\mathbf{p}$, generated by the implicit differential renderer, and the ground-truth color $\mathbf{c}^{\text{gt}}_{\mathbf{p}}$ at pixel $\mathbf{p}$ from the input image: $L_{\text{color}} = (1/|\text{P}|) \cdot \sum_{\mathbf{p}\in \text{P}}| \hat{\mathbf{c}}_{\mathbf{p}} - \mathbf{c}^{\text{gt}}_{\mathbf{p}}|$. This loss helps to refine the geometry and texture of the implicit representation. We employ the Implicit Differentiable Renderer (IDR)~\cite{yariv2020multiview}, which enables the efficient and differentiable rendering of an implicit model.

Similarly, the normal loss $L_{\text{norm}}$ measures the discrepancy between the predicted rendered normal $\hat{\mathbf{n}}_\mathbf{p}$ at pixel $\mathbf{p}$ and the pseudo ground-truth normal $\mathbf{n}^{\text{gt}}_{\mathbf{p}}$ at pixel $\mathbf{p}$ using the normal map $\mathbf{N}_{\rm gt}$. It can be formulated as follows: $L_\text{norm} =(1/|\text{P}|)\cdot \sum_{\mathbf{p}\in{\text{P}}}\|\hat{\mathbf{n}}_{\text{p}}-\mathbf{n}^{\text{gt}}_{\mathbf{p}}\|_2.$ 

Lastly, the eikonal loss $L_{\text{eik}}$ is a regularization term inspired by IGR~\cite{gropp2020implicit} that enforces the property of signed distance functions, where the L2-norm of the gradient of sampling points on the surface should be 1. This is achieved by computing the L2-norm of the gradient of the sampling points on the surface and penalizing deviations from 1 using a squared loss. \\
The overview of the network for this section is shown on the right side of Fig.~\ref{fig:pipeline_stage1}.

\noindent\textbf{Implementation Details.}
Our network inputs are pose-variant explicit geometry information and appearance features from Sec~\ref{sec:method_explicit} and outputs are motion-dependent implicit geometry and rendered appearance. 

\noindent To handle modality mismatch between the 3D mesh and 2D inputs to the neural network, in practice, we encode the geometry in the UV coordinates, $\mathcal{G}_\text{p}=u^{-1}(\mathbf{G}_\text{p})$, where the inverse UV mapping function $u^{-1}$ converts the 3D geometry $\mathbf{G}_{\text{p}}$ into a 2D representation in the UV domain $u^{-1}:\mathbb{R}^{3}\rightarrow\mathbb{R}^{2}$.
Since the UV coordinate is pose-invariant representation, we represent the motion by simply concatenating temporally consecutive $\mathcal{G}_{\text{p}}$:
$\bigoplus(\mathcal{G}_{\text{p}^{t}}, \mathcal{G}_{\text{p}^{t-1}}, \cdots, \mathcal{G}_{\text{p}^{t-n}})$. Utilizing this, we extract global motion features and local motion features for implicit motion-dependent appearance modeling with a U-Net structure:
\begin{eqnarray}
&& E_{\rm m}: \bigoplus(\mathcal{G}_{\text{p}^{t}}, \mathcal{G}_{\text{p}^{t-1}}, \cdots, \mathcal{G}_{\text{p}^{t-n}}) \in\mathbb{R}^{\text{T}\times128\times128\times3} \;  \nonumber \\
&& \rightarrow \;\mathcal{M}_{t}^{\rm G}\in\mathbb{R}^{1\times64}, \;
\mathcal{M}_{t}^{\rm L}\in\mathbb{R}^{128\times128\times64}
\label{eq:7}
\end{eqnarray}
where the global motion feature $\mathcal{M}_{t}^{\rm G}$ is from the latent space of the network, and the local motion features $\mathcal{M}_{t}^{\rm L}$ is from the final output from the decoder. We assign the extracted motion features to the closest 3D points sampled for the implicit motion-dependent avatar.

For the texture network, we modify the IDR~\cite{yariv2020multiview}. It receives the zero isosurface's point, its normal, view direction, and its global geometry feature and outputs the point's color along the view direction. Furthermore, we modify this network to render our implicit geometry with high-quality texture. Our modified network receives appearance feature $\mathbf{A}_{\text{p}^{t}}$ estimated from Section~\ref{sec:method_explicit} as additional input. We feed $\mathbf{A}_{{\text{p}}}$ to each point after UV unwrapping changing the shape from $128\times128\times64$ to $27,554\times64$, and perform the same local sampling. Using this texture network, we generate the final image $\mathbf{I}_\text{M}$. Please refer to the supplementary document for the network details.  \\
\begin{figure*}
    \centering
    \includegraphics[width=1\linewidth]{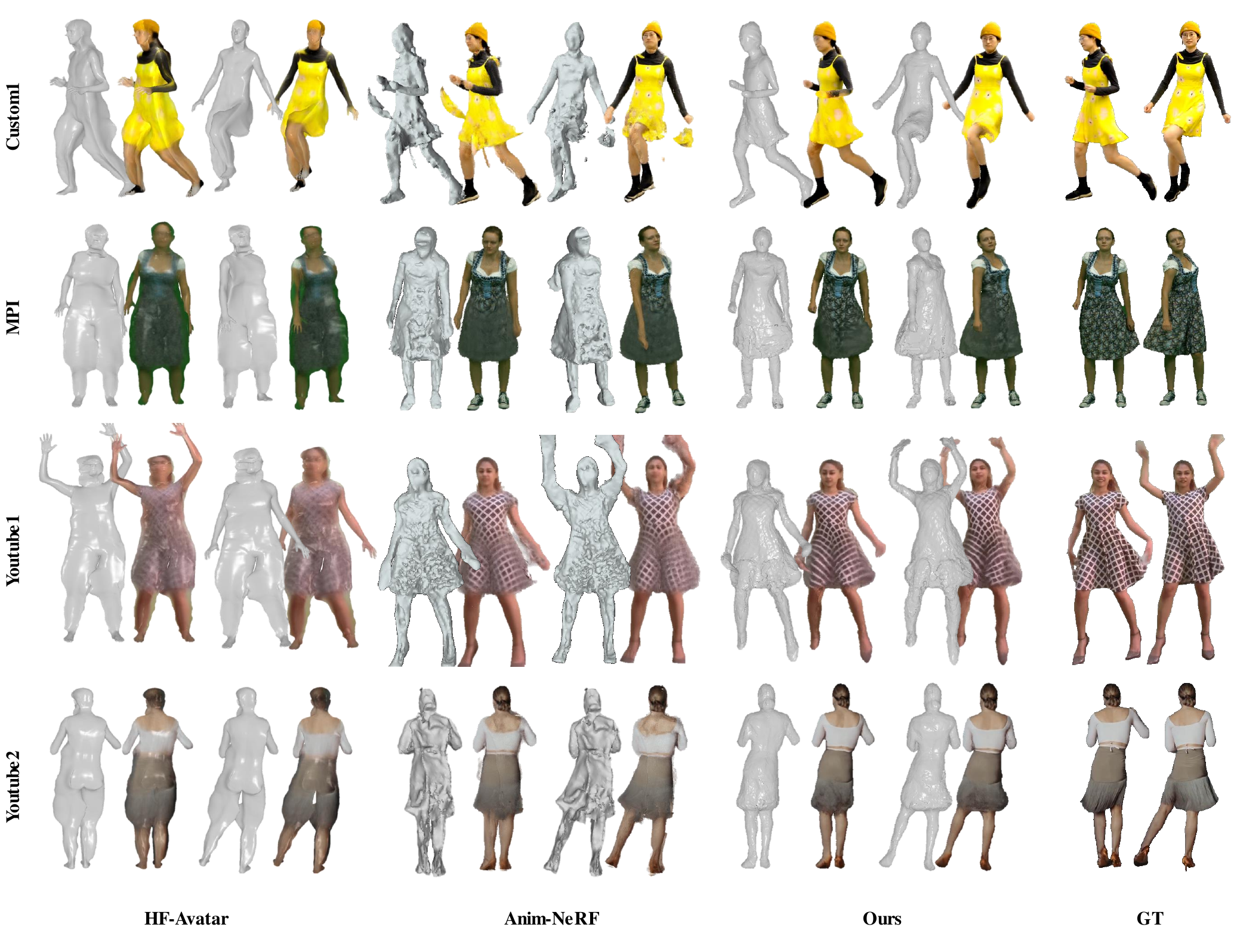}
    \caption{{We compare our method to several baselines (HF-Avatar~\cite{Zhao_2022_CVPR}, Anim-NeRF~\cite{chen2021animatable}) on various sequences(Youtube1-2, MPI and Custom1). For each example, we show the ground truth (GT) target appearances, the rendered appearances by each method, and the reconstructed geometries from sampled two frames.}}
    \label{fig:qualitative}
\end{figure*}

\begin{table*}[h]
    \centering
    \resizebox{\textwidth}{!}{
\begin{tabular}{l|ccc|ccc|ccc|ccc}
\cline{1-13} 
&\multicolumn{3}{c}{Custom 1} & \multicolumn{3}{|c}{MPI} & \multicolumn{3}{|c}{Youtube1} & \multicolumn{3}{|c}{Youtube2} \\
\cline{2-13}
&SSIM$\uparrow$ &LPIPS $\downarrow$ &tOF$\downarrow$ &SSIM $\uparrow$ &LPIPS$\downarrow$ &tOF$\downarrow$ &SSIM $\uparrow$ &LPIPS$\downarrow$ &tOF $\downarrow$ &SSIM $\uparrow$ &LPIPS$\downarrow$ &tOF $\downarrow$ \\
\hline
\textit{HF-Avatar}~\cite{jafarian2021learning} & 0.865 & 0.115& 0.7021 & 0.878 & 0.132 & 0.4861 & 0.836 &0.207 & 1.1018 & 0.8690 & 0.182 & 0.7921\\
\textit{Anim-NeRF}~\cite{chen2021animatable} & 0.888 & 0.109 & 0.6088 & 0.867 & 0.106 &  0.5082 & 0.859 & 0.101 & 0.8573 & 0.8782 & 0.149 & 0.6130\\
Ours & \textbf{0.911} & \textbf{0.074} & \textbf{0.5031} & \textbf{0.882} & \textbf{0.077} & \textbf{0.4220} & \textbf{0.872} & \textbf{0.086} & \textbf{0.7877} & \textbf{0.9122} & \textbf{0.078} & \textbf{0.4176} \\
\hline
\end{tabular}}
\caption{{Quantitative Results. We compared our method to state-of-the-arts (HF-Avatar~\cite{Zhao_2022_CVPR}, Anim-NeRF~\cite{chen2021animatable} on four subjects. We evaluated on three metrics: SSIM $(\uparrow$), LPIPS ($\downarrow$), and tOF ($\downarrow$). Our method outperforms state-of-the-arts for all subjects.}}
\label{table1}
\end{table*}

\begin{figure*}
    \centering
    \includegraphics[width=1\linewidth]{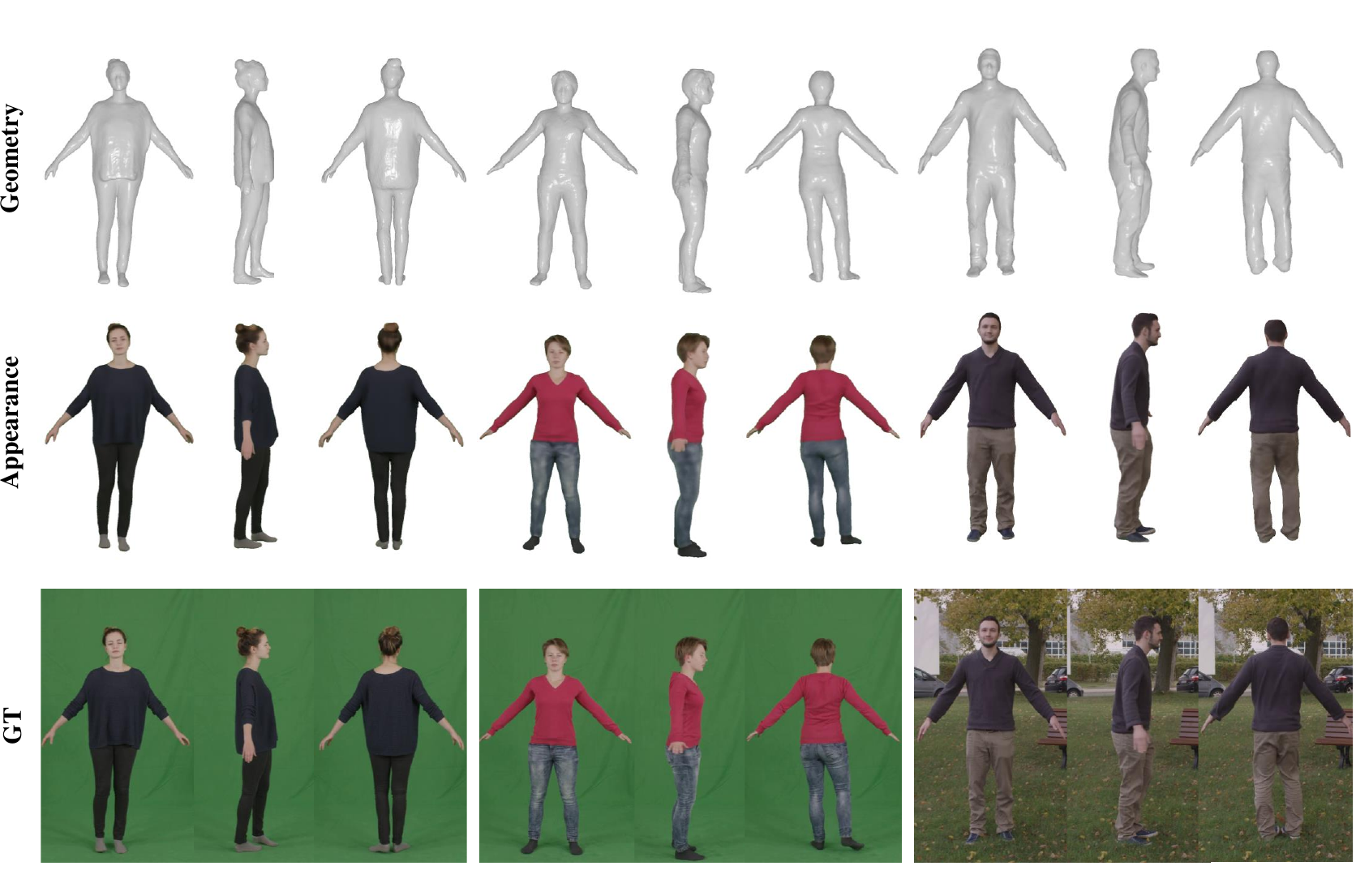}
    \caption{More qualititative results of our method tested on \textit{People-Snapshot}~\cite{alldieck2018video} dataset. This is reconstruction results for ground truth images. Sequential rows depict our method's geometry results, rendered appearance results, and ground truth images (GT). }
    \label{reconstruction}
\end{figure*}

\begin{figure*}
    \centering
    \includegraphics[width=1\linewidth]{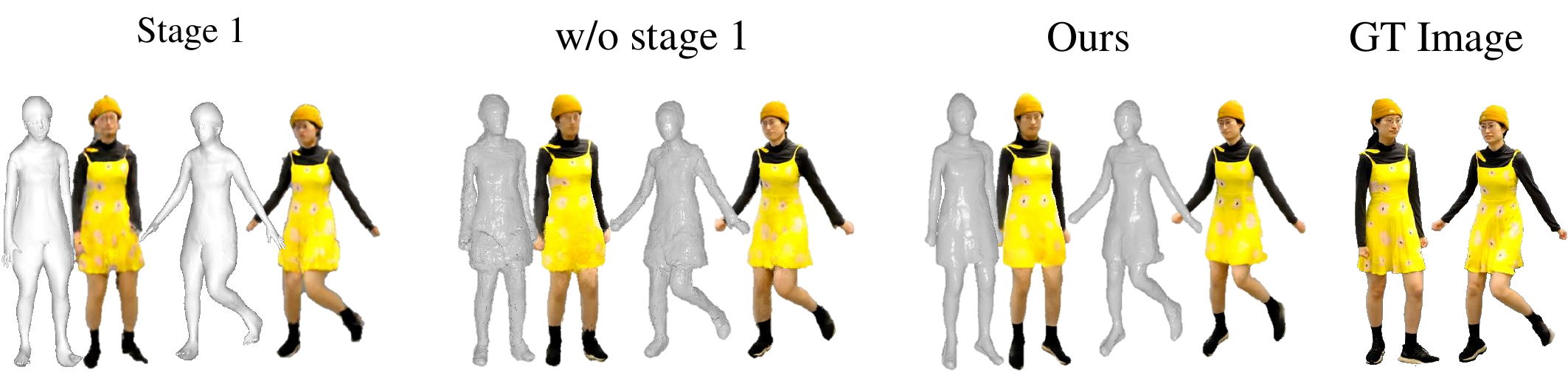}
    \caption{{The ablation study on each module of our method: (column1) Stage 1: only with the explicit pose-dependent appearance model, (column2) without Stage1: utilizing temporal motion features extracted from unclothed human model meshes with the implicit human representation, (column3) Ours.}}
    \label{fig:ablation study}
\end{figure*}

\begin{table}[t]
    \centering
    \begin{tabular}{lccc|}
    \cline{1-4} 
    &\multicolumn{3}{|c}{Custom 1}\\
    \cline{2-4}
    &\multicolumn{1}{|c}{SSIM$\uparrow$} &\multicolumn{1}{c}{LPIPS $\downarrow$} &\multicolumn{1}{c}{tOF $\downarrow$} \\
    \cline{1-4}
    Stage 1 (explicit model) & \multicolumn{1}{|c}{\textbf{0.923}} & \multicolumn{1}{c}{0.092} & \multicolumn{1}{c}{0.5034} \\
    w/o Stage 1 (only SMPL)  & \multicolumn{1}{|c}{0.904} & \multicolumn{1}{c}{0.081} & \multicolumn{1}{c}{0.5240} \\
    Ours & \multicolumn{1}{|c}{0.911} & \multicolumn{1}{c}{\textbf{0.074}} & \multicolumn{1}{c}{\textbf{0.5031}} \\
    \cline{1-4}
    \hline
    \end{tabular}
    \caption{Ablation of each stage. We evaluated its module on Subject 1. Each module shows the significant performance improvement on three metrics: SSIM $(\uparrow$), LPIPS ($\downarrow$), and tOF ($\downarrow$).}
    \label{table2}
\end{table}

\begin{figure*}
    \centering
    \includegraphics[width=1\linewidth]{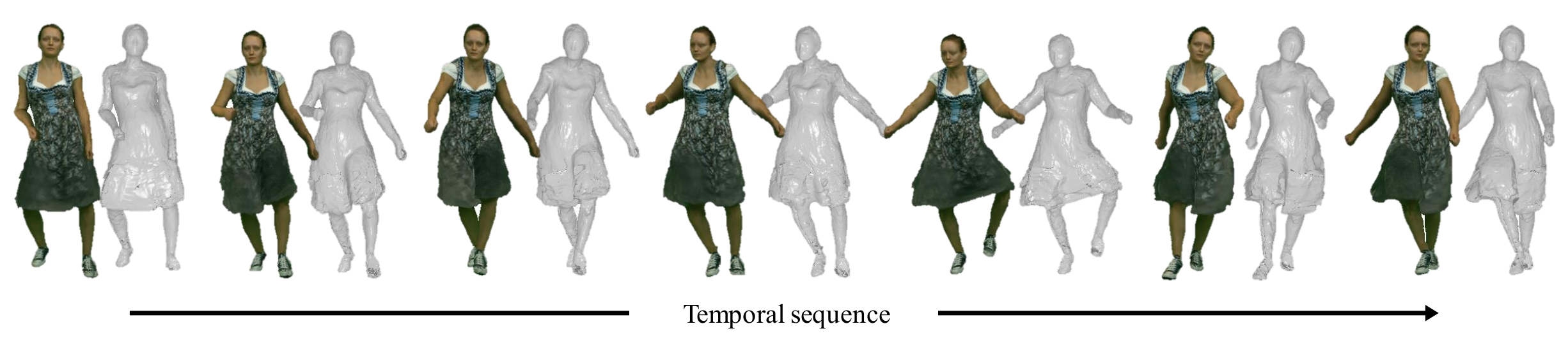}
    \caption{{Our method can generate motion avatars with random SMPL sequences. We utilize SMPL parameters from publicly available AIST~\cite{aist-dance-db} datasets to obtain random parameters for unseen poses, and leverage our previously trained MPI dataset to generate new motion-dependent avatars.}}
    \vspace{1mm}
    \label{application}
\end{figure*}

\section{Experiments}
\label{sec:Experiments}
\noindent\textbf{Datasets.}
We perform experiments using a number of video sequences of people in loose clothing with dynamic movements such as dancing and jumping captured from a previous work~\cite{Yoon_2022_CVPR}, which are including two dance videos from YouTube, and a video sequence from prior work~\cite{kappel2021high}. We use l.5k frames from each video for training and measure the performance on unseen sequences with 200 frames. We use pose and camera parameters extracted from~\cite{kocabas2020vibe} for the test sequences. We also use \textit{People-Snapshot} dataset~\cite{alldieck2018video} for the qualitative result. This dataset provides videos of subjects rotating in the A-pose. Unlike other datasets, the \textit{People-Snapshot} dataset is characterized by small motion dynamics.

\noindent\textbf{Metrics.} We evaluate the quality of rendered images from the reconstructed 3D avatars using three metrics: 1) Structure Similarity (SSIM)\cite{wang2004image} assesses luminance, contrast, and structure quality; higher SSIM indicates greater similarity to the ground truth. 2) Learned Perceptual Image Patch Similarity (LPIPS)\cite{zhang2018unreasonable} compares generated images with ground truth in feature space using a deep neural network, determining perceptual similarity. 3) Temporal Optical Flow (tOF) measures the difference in optical flow between synthesized and ground-truth videos; lower tOF scores suggest more physically plausible appearances relative to real videos.

\noindent\textbf{Baselines.}
In our experiments, we compare our method to three state-of-the-art animatable modeling approaches using a single camera. HF-Avatar~\cite{Zhao_2022_CVPR} refines the unclothed body shape of an explicit 3D model by estimating clothing shape offsets. Anim-NeRF~\cite{chen2021animatable} employs an implicit Neural Radiance Field to model 3D avatars and a skinning function for pose adaptation. For HF-Avatar, we provide initial SMPL pose, shape, and camera parameters extracted from~\cite{kocabas2020vibe} due to failures in our evaluation dataset with significant motion dynamics.


\noindent\textbf{Results.}
We qualitatively and quantitatively evaluate our dynamic human modeling approach against baseline methods. Table~\ref{table1} shows our method outperforms baselines in image quality and temporal plausibility. Fig~\ref{fig:qualitative} demonstrates our method's ability to generate high-quality clothed human avatars conditioned on motion.
HF-Avatar struggles with rotating motions and over-smoothed results due to topology constraints. Anim-Nerf shows significant noise around clothing with large dynamics, failing to learn motion-dependent appearance.
Our method excels in generating high-quality clothed human avatars. Our method captures fine details in geometry and appearance, accurately corresponds motion features with implicit geometry, and employs an adapted texture network to generate high-quality textures, thus outperforming baseline methods in image quality and temporal plausibility. We also demonstrate our avatar modeling results including the geometry and appearance on \textit{People-Snapshot}~\cite{alldieck2018video} data in Fig.~\ref{reconstruction}. By achieving high-quality results on this low-motion dataset, we demonstrate the robust and consistent performance of our approach under a variety of conditions.


\noindent\textbf{Ablation Study.}
Our compositional human modeling pipeline consists of two stages. Stage 1 estimates pose-dependent explicit 3D reconstruction of clothed humans and UV appearance features. In stage 2, implicit networks refine geometry conditioned on motion features on the 3D-clothed surface. Fig.~\ref{fig:ablation study} and Table.~\ref{table2} show our full method outperforms ablation baselines in terms of LPIPS and tOF. The 3D human model generated with explicit representation only (\textit{Stage1}) is constrained by SMPL topologies, limiting fine-grained surface geometry. Using unclothed SMPL body models without Stage 1 (\textit{w/o Stage 1}) restricts fine-grained 3D point sampling, leading to drops in geometric quality. However, \textit{Stage 1} performs better in SSIM measurement as it heavily relies on generative neural network functions without considering geometric quality impact.

\noindent\textbf{Application.}
Our method can generate 3D avatar animation from random SMPL sequences. Fig.~\ref{application} shows an example of its application. We show that our method is highly generalizable to unseen poses. Please refer to the supplementary video for the full animation.

\section{Conclusion}
\label{sec:Conclusion}
In this paper, we introduce a method to model a dynamic clothed 3D human avatar with a plausible motion-dependent appearance using a monocular video. To achieve this, we propose a novel compositional framework that can take the benefits from both explicit and implicit approaches. An explicit network predicts the UV texture features and topology-invariant geometry residuals, and an implicit function constructs high-fidelity local geometry and appearance conditioned on the explicit surface motion. We evaluate our method on multiple video sequences with large motion dyanmics, and our method shows the best performance compared to previous methods. 

\noindent\textbf{Limitation.}
Since our training method is completely free from 3D ground truth, it often includes reconstruction noise on geometry. Our rendering model shows a weak performance (blur artifacts) to learn the appearance with fast motion. Our method highly depends on initial body mesh alignment performance, i.e., the errors from monocular performance tracking propagate to the human avatar modeling pipeline.

{\small
\bibliographystyle{ieee_fullname}
\bibliography{egbib}
}

\end{document}